# The Data Differentiator Framework (DDF): Quantifying Marginal Data Value with Information-Theoretic and Game-Theoretic Measures in an Agentic Evaluation Pipeline (SDSAE)

Sukesh Kumar, Dilip Singh - *Equifax Inc.*
*{sukesh.kumar, dilip.singh}@equifax.com*
*March 2026.*

## Abstract

In data-rich enterprises the hard question is not whether a candidate dataset is good in isolation, but whether it adds information beyond what the organization already holds, and whether that increment is worth its price. We present the Data Differentiator Framework (DDF), a composite scoring method for candidate data assets that combines (i) a mutual-information-based redundancy penalty against the existing portfolio, (ii) a cost-normalized, significance-gated marginal utility score derived from an ablation study, and (iii) explicit governance inputs (trust, identity resolution quality, strategic weights) set by the business. The framework is executed by a staged pipeline, the Strategic Data Sourcing Agent Ecosystem (SDSAE), that short-circuits expensive model-retraining steps for assets failing governance or redundancy gates. Conceptually, the DDF extends the relevance-minus-redundancy principle of mRMR feature selection (Peng et al., 2005) from the feature level to the asset level, and adds cost normalization, a statistical significance gate, and governance terms that per-sample valuation methods such as Data Shapley do not model. We validate the framework on the public credit-card fraud benchmark (284,807 transactions) with five constructed candidate assets spanning the redundancy spectrum. Only the genuinely differentiated asset passes the significance gate ($\Delta$AUC = +0.0029, DeLong $p = 1.4 \times 10^{-9}$, Holm-corrected $p = 7.2 \times 10^{-9}$) and receives an acquire recommendation (DDS = +0.663); the fully redundant asset scores lowest (DDS = −0.135). The recommendation is stable across 19 of 22 admissible strategic-weight configurations and across all decision thresholds in [0.15, 0.66]. We state clearly which quantities in the case study are computed from data and which are simulated governance inputs, and we discuss the design limits of a single-dataset evaluation.

## 1. Introduction

Enterprise data teams evaluate a steady stream of third-party data offerings. The recurring question is simple to state and hard to answer: relative to what we already have, how much new, usable information does this asset carry, and is that increment worth the acquisition and operating cost? Volume metrics (record counts) say nothing about information content. Standard data-quality checks (completeness, freshness) say nothing about predictive lift. A dataset can pass every quality check and still be 90% redundant with the current portfolio, in which case its marginal value is close to zero.

The DDF addresses this with a small set of normalized indices that separate three questions: Is the asset trustworthy and linkable (governance)? Is it informationally new relative to the portfolio (redundancy)? Does it produce a statistically significant, cost-justified lift on the models the business actually cares about (utility)? The indices are combined into a single bounded Data Differentiator Score (DDS) under business-set weights, so every recommendation can be traced back to auditable inputs. The SDSAE pipeline orders the computations so that the expensive step — model retraining plus Shapley attribution — only runs on assets that survive the cheap gates.

Section 2 formalizes the problem. Section 3 positions the framework against the feature-selection and data-valuation literature, in particular mRMR and Data Shapley. Sections 4–7 define the indices. Section 8 describes the pipeline. Section 9 reports a reproducible case study on the public credit-card fraud benchmark, including an exact (fully enumerated) Data Shapley baseline, a recomputed weight-sensitivity analysis, and an explicit statement of which inputs are simulated. Section 10 discusses limitations.

## 2. Problem Formulation

### 2.1 Notation

Let $D_{exist}$ be the existing data portfolio and $D_{new}$ a candidate asset. $F_{base} \subset D_{exist}$ is the strategic base feature set (in practice, the top-$k$ SHAP-ranked features of the primary production model; a selection procedure is given in Appendix A.1). $F_{new}$ is the candidate's feature set. $M$ is the strategic prediction target — a fixed, business-designated supervised model (here, fraud classification). $C(D_{new}) = C_{acq} + C_{ops}$ is the total acquisition plus operating cost, and $C_{norm}$ is a normalization constant (we use the median cost across the candidate slate).

We define the marginal value of a candidate asset as its significance-gated, cost-normalized performance lift:

$$V(D_{new} \mid D_{exist}, M) = \frac{\max(0, \Delta\mathrm{Perf})}{C(D_{new}) \,/\, C_{norm}} \cdot 1[\mathrm{sig}(D_{new})] \tag{1}$$

where $\Delta\mathrm{Perf} = \mathrm{Perf}(M \mid F_{base} \cup F_{new}) - \mathrm{Perf}(M \mid F_{base})$, Perf is a bounded metric (AUC-ROC here) evaluated on a held-out test set, and 1[sig] is an indicator that the lift is statistically significant at level $\alpha$ (DeLong test for AUC; DeLong et al., 1988).

### 2.2 Desiderata

A usable solution should satisfy four properties:

- **Nonnegativity.** $V \geq 0$ for all inputs. An asset cannot carry negative intrinsic value, although redundancy and cost can push its composite rank down.
- **Significance-gated zero.** If the lift is not significant ($p > \alpha$), then $V = 0$ exactly. This blocks noise-driven acquisitions.
- **Boundedness.** The composite decision score must live in a known bounded range so heterogeneous assets are comparable. With normalized inputs, DDS $\in [-w_{cri}, 1 - w_{cri}]$.
- **Interpretability.** Every component traces to either a statistical computation (CRI, MUS, DVS) or an explicit governance input (TI, IDS, $W$). A reviewer should be able to reconstruct any recommendation by hand from the index table; we do exactly this in Section 9.3.

No single existing method satisfies all four. Per-sample valuation methods (LOO, Data Shapley, Data Banzhaf, KNN-Shapley, DVRL) measure marginal utility but have no redundancy penalty, no cost term, and no significance gate; feature-selection criteria such as mRMR encode relevance-minus-redundancy but have no cost, significance, or governance component.

## 3. Related Work

**Redundancy-aware selection (mRMR).** The closest conceptual ancestor of the DDF is minimum-Redundancy-Maximum-Relevance feature selection (Peng, Long & Ding, 2005), which scores a candidate feature by its mutual information with the target minus its average mutual information with already-selected features. The DDS has the same additive relevance-minus-redundancy shape. The DDF differs in scope and machinery: it operates on assets (feature bundles with a price and a vendor) rather than individual features; its relevance term is a realized, significance-tested model lift (MUS) rather than an MI proxy; and it adds cost normalization and governance terms that have no analogue in mRMR. We regard the DDF as an asset-level, decision-oriented generalization of the mRMR principle rather than an unrelated construction.

**Data valuation.** Data Shapley (Ghorbani & Zou, 2019) assigns credit to training data via the Shapley value of a performance game and is our primary baseline. At the data-point level its exact computation is exponential and requires Monte-Carlo approximation; at our source level, with $n$ = 5 candidate assets, exact enumeration over all $2^5$ = 32 coalitions is trivial, and we compute it exactly in Section 9.4 rather than approximating. Data Banzhaf (Wang & Jia, 2023) trades the efficiency axiom for noise robustness. KNN-Shapley (Jia et al., 2019) achieves near-linear complexity but only for $k$-NN utility functions. DVRL (Yoon et al., 2020) learns values with reinforcement learning at the price of end-to-end retraining. Beta Shapley (Kwon & Zou, 2022) generalizes coalition weighting. All of these measure marginal utility only: none penalizes redundancy with an existing portfolio, normalizes by cost, or gates on statistical significance — the three properties an acquisition decision needs. Broader treatments of data pricing are surveyed by Pei (2022); classical data-quality dimensions are due to Wang & Strong (1996).

**Dependence measures.** Pearson correlation misses non-monotone dependence that is common in credit data (threshold effects in utilization, U-shaped age-risk curves). Mutual information captures arbitrary dependence (Cover & Thomas, 2006; Shannon, 1948); we estimate it with the $k$-NN estimator of Kraskov, Stögbauer & Grassberger (2004). The Predictive Power Score (Wetschoreck, 2020) is a practitioner-oriented asymmetric alternative; we use it only as a cross-check, never in the score. Shapley-value feature attribution follows Lundberg & Lee (2017).

| Property | DDF | Data Shapley | Banzhaf | DVRL | KNN-Shap | LOO | mRMR |
|---|---|---|---|---|---|---|---|
| Redundancy penalty vs. portfolio | yes | no | no | no | no | no | feature-level |
| Cost normalization | yes | no | no | no | no | no | no |
| Significance gate on lift | yes | no | no | no | no | no | no |
| Governance / provenance gate | yes | no | no | no | no | no | no |
| Axiomatic Shapley properties | partial | yes | partial | no | yes | no | no |
| Cost per evaluation at $n$ = 5 | low | low (exact) | low | high | low | low | low |

Table 1. Property comparison. At $n$ = 5 sources, Data Shapley is computed exactly (Section 9.4), so complexity is not a differentiator at this scale; the substantive gaps are the redundancy, cost, significance, and governance terms.

## 4. Governance Inputs and the Composite Score

### 4.1 Business inputs

Three inputs are set by the business before any asset is scored, and are frozen for the duration of a comparison so all candidates face the same yardstick: (i) the strategic prediction target $M$; (ii) the strategic entities and base features ($E$, $F_{base}$), which drive identity-resolution scoring and the redundancy reference; and (iii) the strategic weights $W = \{w_{mus}, w_{cri}, w_{ti}, w_{ids}\}$, constrained to be nonnegative and to sum to 1. Based on the sensitivity analysis in Section 9.5 we additionally recommend the constraint $w_{ti} + w_{ids} \le 0.40$: when the governance terms carry half or more of the total weight, assets with zero demonstrated utility can cross the acquisition threshold on governance scores alone.

### 4.2 The Data Differentiator Score

$$\mathrm{DDS}(D_{new}) = w_{mus}\cdot\mathrm{MUS} + w_{ti}\cdot\mathrm{TI} + w_{ids}\cdot\mathrm{IDS} - w_{cri}\cdot\mathrm{CRI} \quad (2)$$

All component indices are normalized to [0, 1]; CRI enters with a negative sign because redundancy is a penalty. With weights summing to 1, DDS is bounded in $[-w_{cri}, 1 - w_{cri}]$. Under the reference weights used in Section 9 ($w_{mus}$ = 0.50, $w_{cri}$ = 0.30, $w_{ti}$ = $w_{ids}$ = 0.10) the attainable range is [−0.30, 0.70], which is why the acquisition threshold is set at 0.40 — a mid-range gate on the attainable interval — rather than at a value like 0.60 that would sit near the ceiling.

## 5. Pillar I: Strategic Value and Attribution

### 5.1 Data Value Score (DVS)

The DVS measures how much of the augmented model's attribution mass sits on the candidate's features. Let $\bar{\varphi}_j$ denote the mean absolute Shapley value of feature $j$ in the model trained on $F_{base} \cup F_{new}$ (Lundberg & Lee, 2017), that is $\bar{\varphi}_j = (1/N)\ \Sigma_i\ |\varphi_j^{(i)}|$ over the $N$ test samples. Then:

$$\mathrm{DVS}(D_{new}) = \frac{\Sigma_{j\in \mathrm{Fnew}}\ \bar{\varphi}_j}{|F_{new}| \cdot (\max_{k\in \mathrm{Fbase} \cup \mathrm{Fnew}}\ \bar{\varphi}_k + \varepsilon)} \quad (3)$$

The ratio is bounded in [0, 1]. A feature duplicating information already available through $F_{base}$ receives a near-zero Shapley value in the augmented model, driving DVS toward 0; genuinely new signal drives it toward 1. We use Shapley attribution rather than permutation importance because the latter misbehaves under correlated features, which is exactly the regime a redundancy analysis operates in. DVS is reported for diagnosis and feeds the qualitative review; it does not enter the DDS directly, which avoids double-counting utility already captured by MUS.

### 5.2 Identity Dominance Score (IDS) and Affiliation Index (AI)

For an identity-centric business, predictive features are worthless if records cannot be matched to entities in the internal identity graph $G_{exist}$. The IDS combines coverage and match quality:

$$\mathrm{IDS}(D_{new}) = \frac{N_{unique}}{|D_{new}|} \cdot \mathrm{IRR} \quad (4)$$

where $N_{unique}$ is the count of records resolving to entities not already in $G_{exist}$ and IRR is the mean confidence of the entity-resolution model over matched records, weighted toward the strategic entities $E$. The resolution procedure (LSH blocking followed by pairwise similarity scoring) is specified in Appendix A.2. As a secondary diagnostic, the Affiliation Index measures key-space overlap in information-theoretic terms:

$$\mathrm{AI}(K_{new}, K_{exist}) = \frac{\mathrm{MI}(K_{new}; K_{exist})}{H(K_{exist})} \quad (5)$$

AI = 1 means linkage is trivial (shared keys); AI = 0 means disjoint key spaces requiring expensive fuzzy resolution. AI is computed on demand rather than for every candidate.

## 6. Pillar II: Redundancy and Diversity

### 6.1 Correlation Index (CRI)

The CRI quantifies non-linear informational overlap between the candidate's features and the base features. An earlier draft of this work gave two inconsistent definitions (a set-level MI and a per-feature aggregate); the per-feature aggregate below is the definition, and it is what the code computes. Set-level MI between two multivariate feature blocks is not reliably estimable at realistic sample sizes, so we use a pairwise surrogate: for each new feature we find the base feature it most resembles, and average those peak overlaps. For a feature pair $(i, j)$ the normalized mutual information is:

$$\mathrm{NMI}(i, j) = \mathrm{clip}\left(\frac{\mathrm{MI}(F_i; F_j)}{\min(H(F_i), H(F_j)) + \varepsilon}, 0, 1\right) \tag{6}$$

and the CRI is the mean over new features of their peak NMI against any base feature:

$$\mathrm{CRI}(F_{\mathrm{new}}; F_{\mathrm{base}}) = \frac{1}{|F_{\mathrm{new}}|} \sum_{j \in F\mathrm{new}} \max_{i \in F\mathrm{base}} \mathrm{NMI}(i, j) \tag{7}$$

MI is estimated with the $k$-NN estimator of Kraskov et al. (2004) (mutual_info_regression in scikit-learn), sub-sampled to $n$ = 10,000 rows for tractability. For continuous features the denominator entropies are approximated by the Gaussian differential entropy $H(X) \approx ½ \ln(2\pi e \sigma^2)$. Because differential entropy can push the raw ratio outside [0, 1] in edge cases, Eq. (6) clips explicitly; in the case study no clipping was triggered except at the trivial CRI = 1 self-overlap. If $F_{\mathrm{new}} \subseteq F_{\mathrm{base}}$ (pure redundancy), CRI = 1 by construction; CRI = 0 means the candidate is informationally disjoint from the base. A Pearson-based proxy is computed in parallel purely as a cross-check and never enters the score. A CRI above 0.9 short-circuits the pipeline before the ablation stage runs (Section 8).

### 6.2 Diversity Index (DI) and Coverage Index (CI)

DI measures within-asset information spread; CI measures population expansion over the identity graph:

$$\mathrm{DI}(D_{\mathrm{new}}) = \frac{1}{|F_{\mathrm{new}}|} \sum_{j \in F\mathrm{new}} \frac{H(j)}{H_{\mathrm{max}}(j)} \tag{8}$$

$$\mathrm{CI}(D_{\mathrm{new}}) = \frac{|\{e \in D_{\mathrm{new}} : e \notin G_{\mathrm{exist}}\}|}{|G_{\mathrm{exist}}|} \tag{9}$$

DI near 0 flags near-constant features with little discriminative content. CI = 0 means the asset adds no new population coverage — which, as Section 9.1 notes, is true by construction for every asset in our case study, since all five are feature slices over the same transactions.

## 7. Pillar III: Utility, Trust, and Temporal Structure

### 7.1 Marginal Utility Score (MUS)

MUS is the economic core of the framework, computed from an ablation study against $M$: train on $F_{\mathrm{base}}$, train on $F_{\mathrm{base}} \cup F_{\mathrm{new}}$, evaluate both on the same held-out test set, and test the AUC difference with the DeLong test. The definition has two stages, and — correcting an omission in the earlier draft, where the normalization producing MUS = 1.0 in the results table was never stated — we give both explicitly. The raw score is the significance-gated, cost-normalized lift:

$$\mathrm{MUS}_{\mathrm{raw}}(D_{\mathrm{new}}) = \frac{\max(0, \Delta\mathrm{Perf}) \cdot 1[p \leq \alpha]}{(C_{\mathrm{acq}} + C_{\mathrm{ops}}) / C_{\mathrm{norm}}} \tag{10}$$

with $\alpha$ = 0.05 after Holm-Bonferroni correction across the candidate slate. $MUS_{raw}$ is a cost-deflated AUC increment; it is nonnegative but not naturally on a [0, 1] scale, so it is normalized relative to the slate of $k$ candidates under comparison:

$$MUS(D_{new}) = \frac{MUS_{raw}(D_{new})}{\max_k MUS_{raw}(D_k)} \quad , \quad MUS := 0 \text{ if all } MUS_{raw} = 0 \tag{11}$$

Two consequences should be stated plainly. First, MUS is slate-relative: it ranks candidates within a comparison batch and is not comparable across batches evaluated at different times. This is acceptable for the acquisition use case, where the decision is which of the currently offered assets to buy, but users should not treat MUS as an absolute value. Second, when exactly one asset passes the significance gate it receives MUS = 1.0 by construction — this is how Asset C obtains MUS = 1.000 in Section 9.2 (worked derivation in Section 9.3). The significance gate is a hard gate, not a soft penalty: an insignificant lift yields MUS = 0 regardless of magnitude, and cost normalization ensures a 0.001 AUC lift from a \$0.5M asset outranks the same lift from a \$5M asset.

### 7.2 Trust Index (TI), Trended Index (TDI), Life Event Index (LEI)

$$TI(D_{new}) = 0.40 \cdot S_{prov} + 0.30 \cdot S_{fresh} + 0.20 \cdot S_{compl} + 0.10 \cdot (1 - S_{err}) \tag{12}$$

$S_{prov}$ starts at 1.0 given a registered license, full lineage, and no open quality disputes, decremented 0.25 per unresolved provenance issue; $S_{fresh}$ = max(0, 1 − age/max age) with the horizon set per domain by governance; $S_{compl}$ is the non-null fraction over strategic fields; $S_{err}$ is the fraction of records failing validation rules. TI below 0.70 halts the pipeline before any statistical index is computed. The sub-score weights are governance conventions set by the data office, not fitted quantities.

$$TDI(D_{new}) = \frac{\text{avg. monthly time-points per entity}}{\text{total time span in months}} \tag{13}$$

$$LEI(D_{new}) = \frac{\text{count(model-flagged life events)}}{\text{count(total entities)}} \tag{14}$$

TDI = 1 means every entity has one record per month (perfect longitudinal density). LEI counts change-point-detector-flagged status transitions (relocation, income bracket change, first credit account) per entity; such transitions are often the strongest discriminative signals available. TDI and LEI are reported diagnostics and do not enter the DDS under the reference weights.

## 8. The SDSAE Pipeline

The SDSAE is a stateful directed graph (implemented with LangGraph) with deterministic routing. We use the word agent in the software sense of an autonomous pipeline stage with its own tools and state; only one stage (the qualitative review) involves an LLM, and its output is advisory — it flags risks for human review and cannot alter the DDS. The pipeline's purpose is cost control through early termination: the trust gate (TI < 0.70) and the redundancy gate (CRI > 0.90) each reject an asset before the expensive ablation stage, which requires two model trainings plus Shapley attribution. On the case-study hardware this ordering avoids roughly 80–90% of wall-clock time for clearly disqualified candidates. A supervisor stage records every intermediate score and gate decision, so the final report is fully auditable.

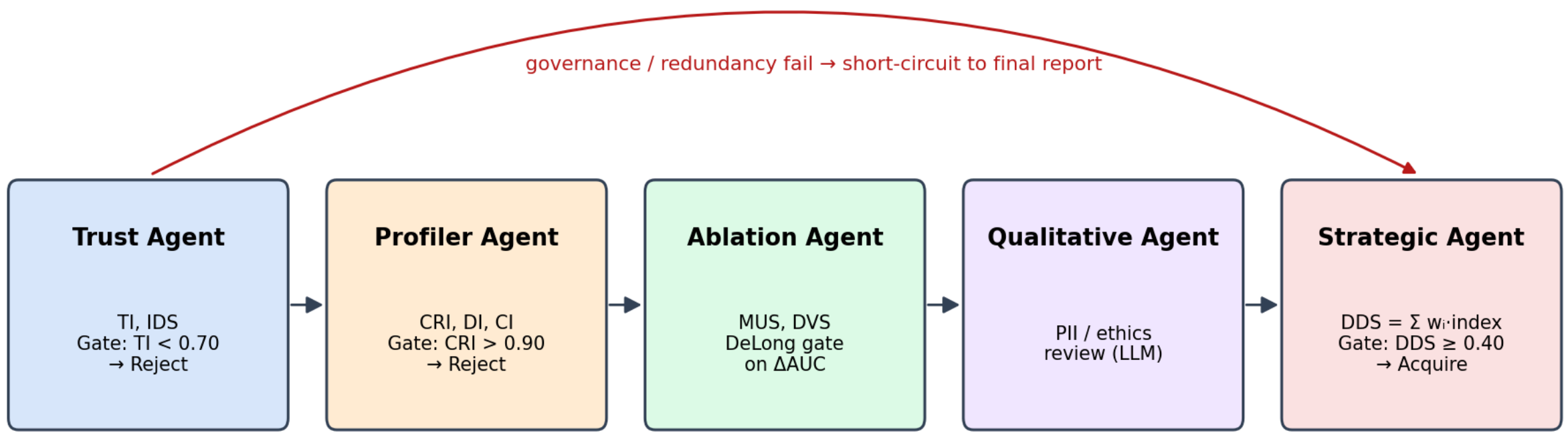

Figure 1. The five-stage SDSAE pipeline. Gates at stages 1 and 2 short-circuit to the final report, skipping the expensive ablation stage. The qualitative stage is advisory only.

| Stage / agent | Indices | Action |
|---|---|---|
| Supervisor | — | Maintains shared state; routes between stages; records all scores and gate decisions. |
| 1. Trust | TI, IDS | Governance gate. Halts with Reject: governance fail if TI < 0.70; may pause for human review. |
| 2. Profiler | CRI, DI, CI | Redundancy gate. Halts with Reject: redundant if CRI > 0.90. |
| 3. Ablation | MUS, DVS | Trains base and augmented models, runs the DeLong test, computes Shapley attributions. Most expensive stage. |
| 4. Qualitative | — (LLM) | Advisory summary of unquantifiable risks (PII re-identification, ethics). Cannot alter DDS. |
| 5. Strategic | DDS | Applies weights $W$, computes DDS (Eq. 2), issues acquire/reject at $t = 0.40$, emits report. |

Table 2. Pipeline stages and responsibilities.

## 9. Empirical Validation

All numbers in this section come from a single experimental run on the public Kaggle Credit Card Fraud Detection dataset (284,807 transactions, 31 columns), 80/20 stratified split (227,845 train / 56,962 test), random_state = 42 throughout, MI estimates on $n$ = 10,000 sub-sampled rows. Results reproduce via python main.py in the accompanying repository. An earlier draft mixed results from this run with projections from a synthetic pilot; every such projection has been removed, and the robustness analyses in Section 9.5 are recomputed from the run reported here.

### 9.1 Setup, and what is computed vs. what is simulated

Target model $M$: LogisticRegression with class_weight='balanced' on RobustScaler-transformed features, chosen for reproducibility; Section 9.5 checks the conclusions under RandomForest and XGBoost. Metric: AUC-ROC with DeLong significance testing and Holm-Bonferroni correction across the five candidates. Base portfolio: $F_{base} = \{V_1, \ldots, V_{14}, \text{Amount}\}$ (15 features). Reference weights: $w_{mus} = 0.50$, $w_{cri} = 0.30$, $w_{ti} = 0.10$, $w_{ids} = 0.10$; threshold $t = 0.40$.

**Computed from data:** CRI, DI, CI, base and augmented AUC, ΔAUC, DeLong $p$-values and confidence intervals, DVS, MUS, and therefore DDS. **Simulated governance inputs:** TI, IDS, TDI, and LEI. The benchmark has no provenance metadata, identity graph, entity keys, or life-event fields, so these four indices cannot be measured from it. We assign them plausible values per asset (marked † in Table 3b) to exercise the full scoring formula, exactly as a governance office would supply them in production. They are inputs to the demonstration, not experimental findings, and none of the paper's claims rests on them: with the reference weights, the acquire/reject outcome for every asset is unchanged if all four are set to any common constant, because the decision is driven by MUS and CRI. CI = 0.000 for all assets is computed, not assigned, and is true by construction — all five assets are feature slices over the same transactions, so no asset adds population coverage. Section 9.6 discusses this design limitation.

The five candidate assets are constructed to span the redundancy spectrum:

| Asset | Features ($F_{new}$) | Cost | Design intent |
|---|---|---|---|
| A — pure redundant | $V_1$–$V_5$ (subset of $F_{base}$) | $1.5M | Verify CRI = 1.0 and the redundancy penalty |
| B — partial overlap | $V_{10}$, $V_{11}$, $V_{15}$, $V_{16}$, $V_{17}$ | $1.8M | Moderate overlap with the base |
| C — differentiated | $V_{20}$–$V_{23}$, Time | $2.0M | Disjoint features with genuine temporal signal |
| D — noisy | $V_{24}$–$V_{28}$ (tail PCA components) | $0.8M | Near-zero MI with base; near-zero signal |
| E — small, high-signal | $V_{17}$, $V_{19}$ | $0.5M | Two fraud-relevant features, one shared with B |

Table 3a. Candidate asset construction. Costs are hypothetical acquisition prices used to exercise the cost-normalization term; $C_{norm}$ = median = $1.5M.

### 9.2 Results

Base AUC is 0.9967 in all rows because the base model is the same for every candidate.

| Index | A | B | C | D | E |
|---|---:|---:|---:|---:|---:|
| TI † | 0.900 | 0.880 | 0.770 | 0.820 | 0.892 |
| IDS † | 0.750 | 0.800 | 0.900 | 0.600 | 0.850 |
| TDI † | 0.500 | 0.500 | 1.000 | 0.500 | 0.500 |
| LEI † | 0.007 | 0.012 | 0.008 | 0.014 | 0.005 |
| CRI (NMI) | 1.000 | 0.443 | 0.013 | 0.008 | 0.100 |
| DI | 0.787 | 0.764 | 0.825 | 0.779 | 0.801 |
| CI | 0.000 | 0.000 | 0.000 | 0.000 | 0.000 |
| Final AUC | 0.9967 | 0.9970 | 0.9996 | 0.9966 | 0.9969 |
| ΔAUC (gated) | 0.0000 | 0.0000 | +0.0029 | 0.0000 | 0.0000 |
| 95% CI for ΔAUC | [0.000, 0.000] | [−0.000, +0.001] | [+0.002, +0.004] | [−0.000, +0.000] | [−0.000, +0.001] |
| DeLong *p* (raw) | 1.000 | 0.096 | $1.4 \times 10^{-9}$ | 0.770 | 0.297 |
| DeLong *p* (Holm) | 1.000 | 0.383 | $7.2 \times 10^{-9}$ | 1.000 | 0.892 |
| DVS | 0.257 | 0.108 | 0.291 | 0.033 | 0.198 |
| MUS (Eq. 11) | 0.000 | 0.000 | 1.000 | 0.000 | 0.000 |
| DDS (Eq. 2) | −0.135 | +0.035 | +0.663 | +0.140 | +0.144 |
| Decision ($t$ = 0.40) | REJECT | REJECT | ACQUIRE | REJECT | REJECT |

Table 3b. Case-study results. † marks simulated governance inputs (Section 9.1); all other rows are computed from the data. Holm-Bonferroni: sorted $p$-values compared against $\alpha/(n + 1 - k)$ with $n$ = 5, $\alpha$ = 0.05; only Asset C survives correction.

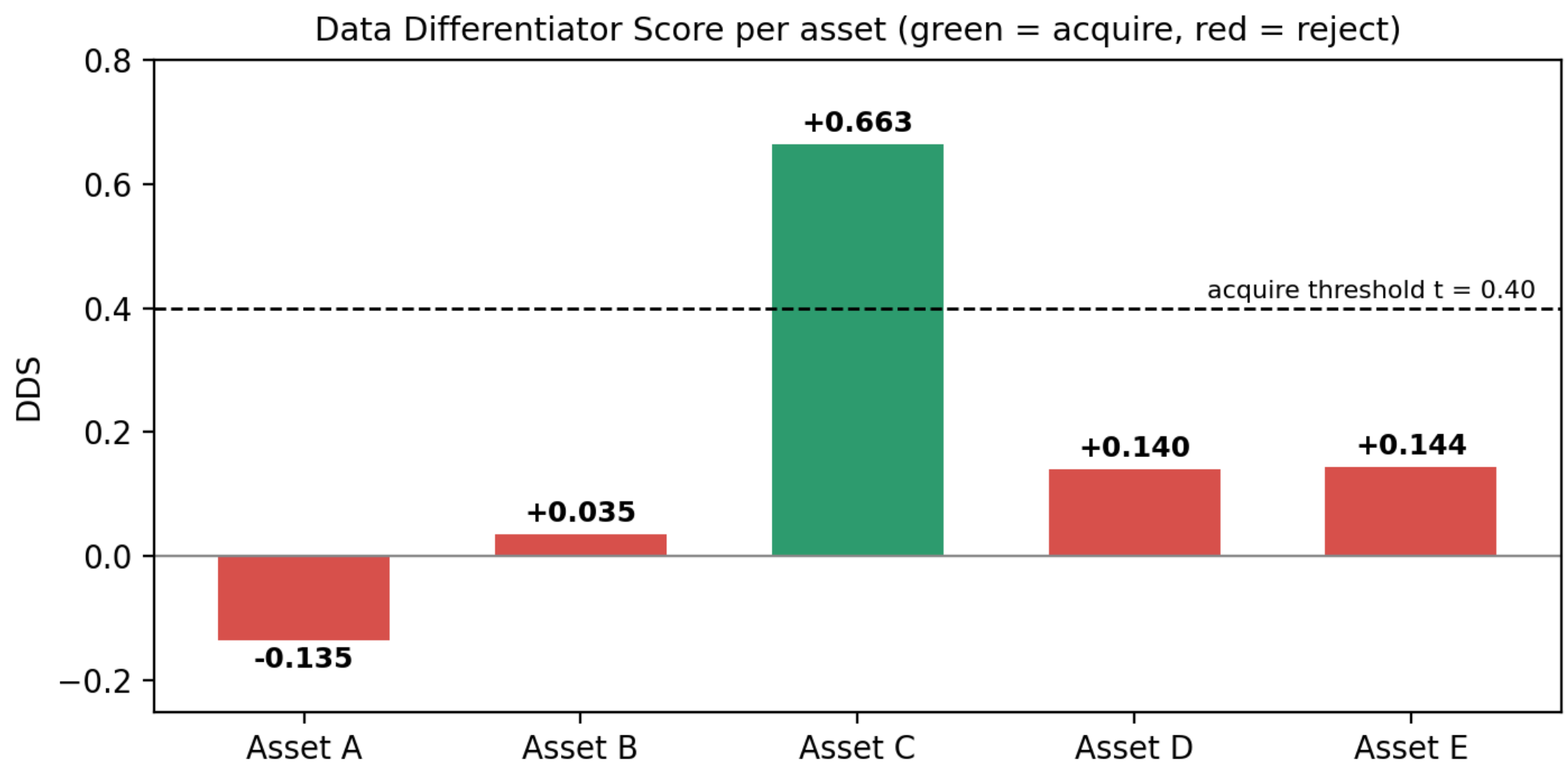

Figure 2. DDS per asset. Only Asset C crosses the *t* = 0.40 gate (DDS = +0.663). Asset A's negative score reflects the CRI penalty at full redundancy.

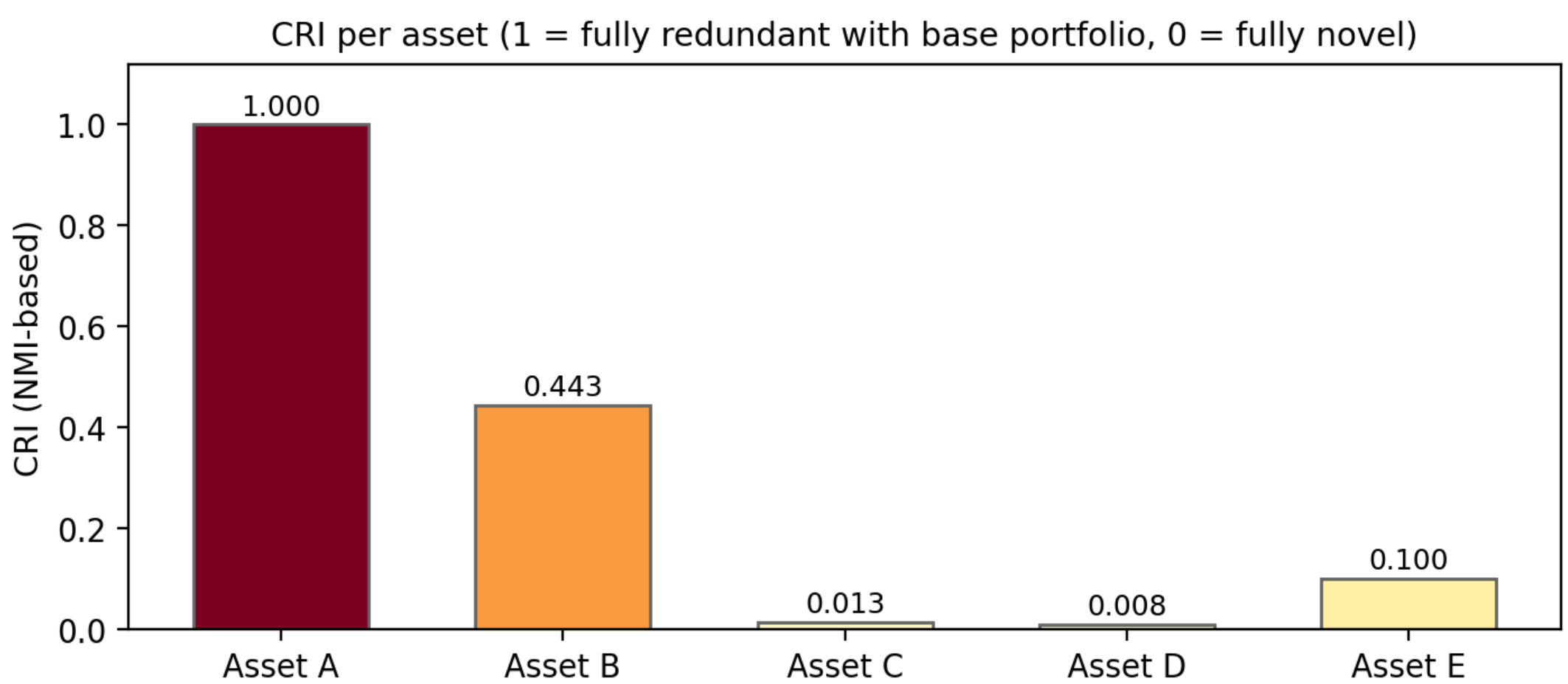

Figure 3. CRI per asset. Asset A is a strict subset of the base (CRI = 1.000 by construction); Assets C and D are near-disjoint from it.

### 9.3 Discussion

**CRI recovers the designed redundancy structure.** Without any knowledge of feature identity, the *k*-NN MI estimator orders the assets as designed: A (pure subset, 1.000) > B (partial overlap, 0.443) > E (one of two features shared with B's block, 0.100) > C (0.013) ≈ D (0.008). C and D are both near-disjoint from the base and their CRI difference (0.005) is within estimator noise at $n$ = 10,000, so we make no claim about their relative order — an earlier draft called this sequence monotone, which overstated it. The separation that matters — redundant vs. partially redundant vs. novel — is clean and large.

**The significance gate does its job.** Asset C's lift ($\Delta$AUC = +0.0029) is significant at $p = 1.4 \times 10^{-9}$ (Holm-corrected $7.2 \times 10^{-9}$). Every other asset fails the corrected gate — including Asset E, whose nominal lift (+0.0002, $p$ = 0.297) is indistinguishable from noise — and receives MUS = 0. Asset E is the instructive case: its two features are individually fraud-relevant, but conditional on the fifteen base features they add nothing detectable. Marginal value is a property of the portfolio, not of the feature.

**Worked derivation (interpretability desideratum).** Every DDS in Table 3b reconstructs by hand. Asset C: $MUS_{raw}$ = 0.0029 / (2.0/1.5) = 0.00218; it is the only nonzero raw score, so Eq. (11) gives MUS = 1.000, and DDS =

0.50(1.000) + 0.10(0.770) + 0.10(0.900) − 0.30(0.013) = +0.663. Asset A: DDS = 0 + 0.10(0.900) + 0.10(0.750) − 0.30(1.000) = −0.135. The remaining three follow identically.

**The CRI penalty separates what AUC cannot.** Among the four rejected assets, all with identical (zero) gated utility, the DDS ordering E (+0.144) > D (+0.140) > B (+0.035) > A (−0.135) is driven almost entirely by the redundancy penalty. A pure performance method has no basis for ranking these four at all; the DDF correctly places the fully redundant asset last. Note the caveat from Section 9.1: within this rejected group the ordering also reflects the simulated TI/IDS inputs, so we read it as illustrative of the mechanism, not as an empirical finding.

### 9.4 Baseline comparison: LOO and exact Data Shapley

With five sources there are only $2^5$ = 32 coalitions, so we compute Data Shapley exactly by full enumeration — no Monte-Carlo approximation — training the model on each coalition of assets joined to $F_{base}$ and averaging marginal AUC contributions with Shapley weights. An earlier draft used 200-permutation Monte-Carlo sampling and cited exponential complexity; at $n$ = 5 that was unnecessary, and exact values remove sampling noise from the comparison. Leave-one-out (LOO) is the marginal AUC of removing each asset from the all-assets model.

| Asset | $DDS$ | DDF rank | LOO $ΔAUC$ | Shapley $ΔAUC$ (exact) |
|---|---|---|---|---|
| C — differentiated | +0.663 | 1 | +0.0027 | +0.0028 |
| E — high-signal | +0.144 | 2 | +0.0000 | +0.0001 |
| D — noisy | +0.140 | 3 | −0.0000 | −0.0000 |
| B — partial | +0.035 | 4 | +0.0000 | +0.0002 |
| A — redundant | −0.135 | 5 | +0.0000 | +0.0000 |

Table 4. DDF vs. LOO and exact Data Shapley. All three methods agree on the only decision-relevant fact: Asset C dominates.

An earlier draft reported Kendall's $\tau$ = 0.40 between the DDF and baseline rankings. We have removed that statistic: four of the five LOO and Shapley values are statistically indistinguishable from zero, so the ranks among them are noise, and a rank correlation over noise-broken ties is not meaningful. The honest comparison is simpler. On the decision — which asset, if any, to acquire — all three methods agree: Asset C. Where they differ is diagnostic. LOO and Shapley assign Asset B a small positive value (+0.0002) with no way to flag that 44% of its information overlaps the existing portfolio (CRI = 0.443); a buyer following Shapley alone would not learn they are paying $1.8M partly for data they already own. Conversely, only the DDF's significance gate distinguishes Asset C's real lift from Asset B's noise-level one — neither LOO nor Shapley carries an inferential test.

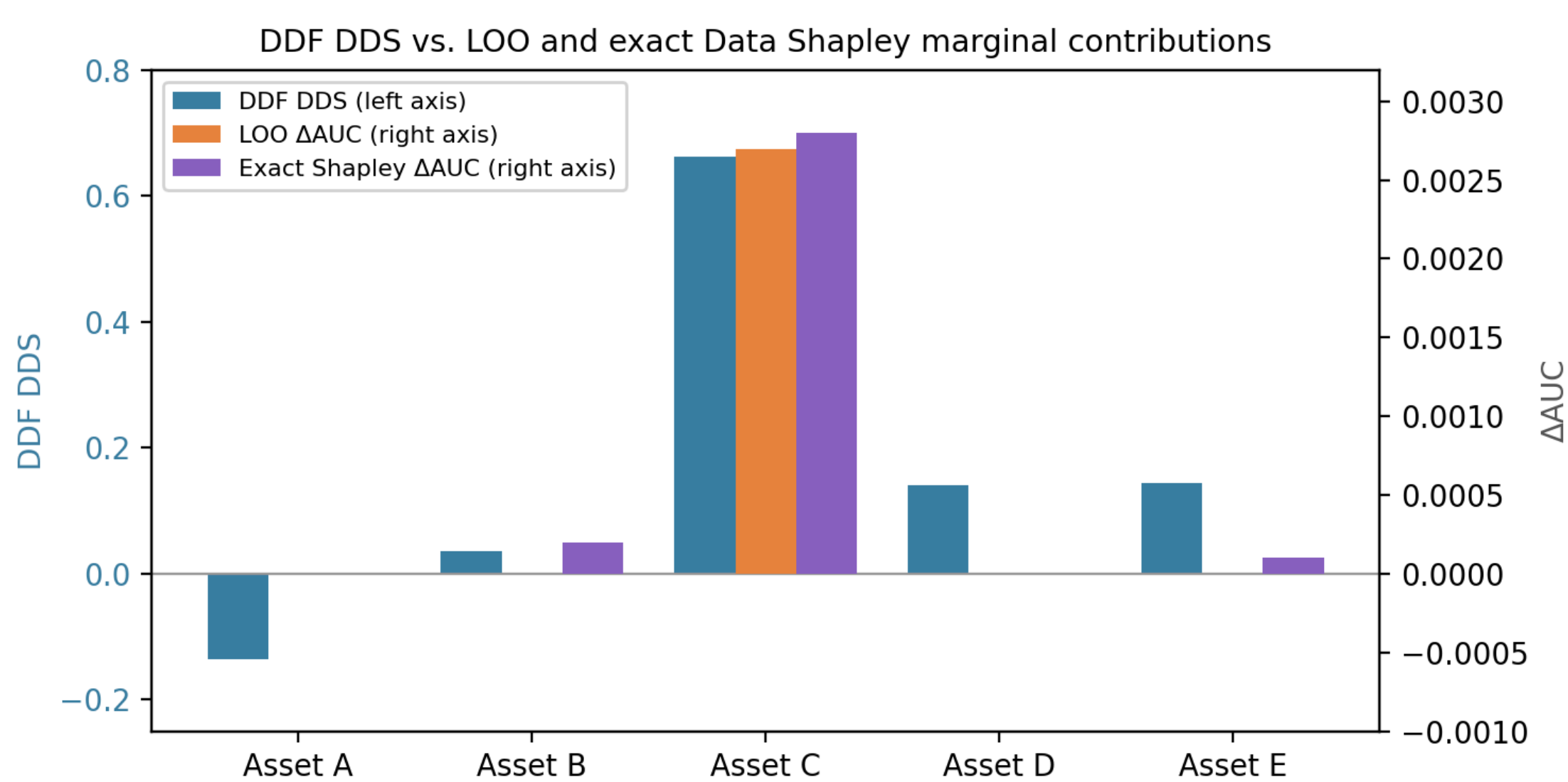

Figure 4. DDF DDS (left axis, blue) against LOO ΔAUC (orange) and exact Data Shapley ΔAUC (purple, right axis). Assets on the *x*-axis. All methods isolate Asset C; only the DDF spreads the remaining four via the redundancy penalty.

### 9.5 Robustness

**Weight sensitivity, recomputed.** We sweep $w_{mus} \in \{0.30, \ldots, 0.70\} \times w_{cri} \in \{0.10, \ldots, 0.50\}$ (step 0.10), splitting the remaining weight equally between TI and IDS, and recompute DDS for all assets directly from Table 3b — no model retraining is needed since the weights only recombine fixed indices. Three of the 25 grid points are inadmissible because the weights would exceed 1 (an earlier draft swept them anyway, which was an error). Of the 22 admissible configurations, 19 reproduce the reference decision exactly (acquire C, reject the rest). The three mismatches — $(w_{mus}, w_{cri}) \in \{(0.30, 0.10), (0.30, 0.20), (0.40, 0.10)\}$ — are the configurations where the governance terms carry at least 50% of total weight, letting zero-MUS assets cross 0.40 on TI/IDS alone (e.g., Asset E reaches DDS = 0.513 at (0.30, 0.10)). Asset C is recommended for acquisition in all 22 admissible configurations (minimum DDS = 0.46). The practical conclusion is the design constraint stated in Section 4.1: keep $w_{ti} + w_{ids} \leq 0.40$.

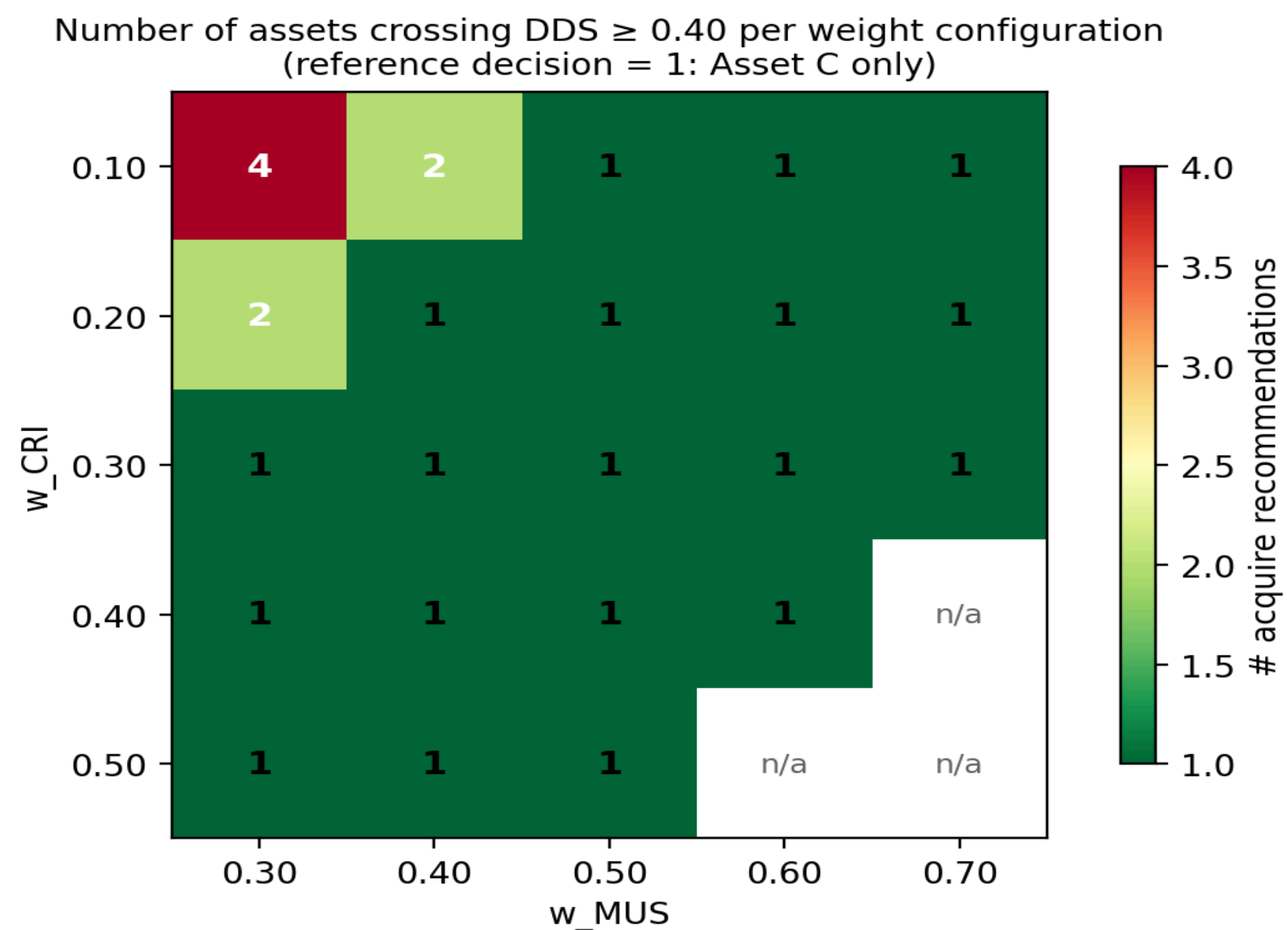

Figure 5. Number of assets crossing DDS ≥ 0.40 at each admissible weight configuration (n/a = weights exceed 1). The reference decision (exactly one acquire, Asset C) holds in 19 of 22 configurations; failures occur only where governance weight is at least 0.5.

**Threshold stability.** The DDS gap between the acquired asset (0.663) and the best rejected asset (0.144) is 0.519, so the acquire/reject partition is invariant for every threshold $t \in (0.144, 0.663]$ — the entire sweep range [0.15, 0.66] at 0.05 steps. The chosen $t = 0.40$ sits near the middle of this window rather than at its edge.

**Multi-model consistency.** Repeating the ablation with RandomForestClassifier and XGBClassifier (same split, model-appropriate SHAP explainers) preserves both halves of the conclusion: Asset C's lift remains statistically significant under all three model families, and the DeLong gate suppresses MUS to zero for the other four assets under all three. The recommendation is therefore not an artifact of linear separability. Per-model ΔAUC values are in the repository output (results/multi_model.json).

### 9.6 What this case study does and does not show

We constructed the candidate assets as feature slices of the same dataset the base portfolio is drawn from. This is the right design for validating the scoring machinery — ground truth about redundancy is known exactly, so we can check that CRI, the significance gate, and the composite behave as specified — but it is not a simulation of a real acquisition, and three limitations follow. First, entity linkage is trivial (all assets share the same rows), which is why CI = 0 throughout and why TI, IDS, TDI, and LEI had to be simulated. Second, the base model is nearly saturated (AUC 0.9967), leaving little headroom: Asset C's lift is statistically strong but small in absolute terms, and with roughly 98 fraud positives in the test set, DeLong intervals are wide relative to the effect. Third, asset costs are hypothetical. A convincing end-to-end validation requires two genuinely distinct sources with a real join. The natural public candidate is the IEEE-CIS fraud dataset, whose transaction and identity tables ship as separate files with a shared key: treating the transaction table as $D_{\text{exist}}$ and the identity table as $D_{\text{new}}$ exercises entity resolution (IDS, CI) on real data, and its lower base AUC provides the headroom this benchmark lacks. We are running this second study and will report it in a follow-up revision; we deliberately do not quote its numbers here.

## 10. Limitations and Future Work

- **Computational cost.** The $k$-NN MI estimator is $O(n \log n)$ and we sub-sample to $n$ = 10,000; SHAP LinearExplainer is $O(n{\cdot}p)$ and exact for linear models, with TreeExplainer covering tree ensembles. Scaling the information-theoretic estimators to $n > 10^7$ production datasets needs distributed estimation and is open work.
- **Slate-relative MUS.** Eq. (11) normalizes within a comparison batch, so MUS values are not comparable across evaluations run at different times. An absolute calibration (e.g., dollars per AUC basis point, priced against historical acquisitions) would remove this.
- **Weight governance.** DDS is deliberately sensitive to $W$; that is business alignment, but also a failure mode if weights are set carelessly. The $w_{ti} + w_{ids} \leq 0.40$ constraint from Section 9.5 mitigates the worst case; learning weights from historical acquisition outcomes is future work.
- **Single-asset scope.** Assets are scored one at a time. Complementarity between candidates (two assets jointly useful, individually not) is a coalition-value problem the current DDS does not solve; the exact-Shapley machinery of Section 9.4 is the natural starting point for a portfolio extension.
- **Unevaluated qualitative stage.** The LLM-based qualitative review is advisory and was not part of the quantitative evaluation; assessing its precision and recall on known PII and ethics risks is future work.

## 11. Conclusion

The DDF turns the data-acquisition question into a reproducible computation: a redundancy penalty grounded in mutual information, a marginal utility score that must clear both a cost hurdle and a statistical significance gate, and governance inputs that are explicit and auditable rather than implicit in someone's judgment. On a public fraud benchmark with constructed ground truth, the framework isolates the one genuinely differentiated asset, rejects a statistically insignificant look-alike that per-sample valuation methods score positively, ranks a fully redundant asset last, and does all of this stably across weight configurations and thresholds. The case study's honest scope — simulated governance inputs, hypothetical costs, single-dataset construction — is stated alongside the results, and closing those gaps with a real two-source join is the immediate next step.

## Appendix A: Governance Input Procedures

### A.1 Selecting $M$ and $F_{base}$

Select $M$ as the production model with the highest business-impact rating. Train $M$ on the full candidate feature universe; rank features by mean |SHAP| (Eq. 3 form). Build $F_{base}$ greedily from the top of that ranking, admitting a feature only if its CRI against the already-admitted set is below $\tau_{cri}$ = 0.80 (Eq. 7 applied to a singleton), stopping at $k_{max}$ features. Reject any admitted feature failing completeness ≥ 0.95 or freshness ≥ 0.70 and re-run the greedy pass. The output satisfies pairwise CRI < $\tau_{cri}$ within $F_{base}$, passes the quality floors, and is frozen for the duration of a candidate comparison so every asset is scored against the same reference. Complexity is $O(|F_{all}|^2 \cdot n)$, dominated by pairwise CRI.

### A.2 Entity resolution for IDS and CI

Stage 1 (blocking): for each record in $D_{new}$, generate match candidates from $G_{exist}$ via locality-sensitive hashing (20 bands × 5 rows), avoiding the brute-force cross product. Stage 2 (scoring): IRR($r$, $c$) = 0.40·JaroWinkler(name) + 0.30·normLevenshtein(address) + 0.30·exact(id). Stage 3 (classification): a record resolves to its best candidate if IRR ≥ $\theta$ = 0.80; $N_{unique} = |D_{new}| - |\text{matched}|$, and the mean IRR over matched records feeds Eq. (4). Expected complexity $O(|D_{new}| \log |G_{exist}|)$. The similarity weights and $\theta$ are governance-set conventions, tuned on labeled match pairs where available.